\pdfoutput=1

\documentclass[11pt]{article}

\usepackage{emnlp2021}

\usepackage{times}
\usepackage{latexsym}

\usepackage[T1]{fontenc}

\usepackage[utf8]{inputenc}

\usepackage{microtype}

\usepackage{makecell}
\usepackage{xspace}
\usepackage{amsmath,amssymb}
\usepackage{multirow}
\usepackage{booktabs}
\usepackage{graphicx}
\usepackage{eqparbox}

\DeclareSymbolFont{extraup}{U}{zavm}{m}{n}
\DeclareMathSymbol{\varheart}{\mathalpha}{extraup}{86}
\DeclareMathSymbol{\vardiamond}{\mathalpha}{extraup}{87}

\makeatletter
\newcommand\footnoteref[1]{\protected@xdef\@thefnmark{\ref{#1}}\@footnotemark}
\makeatother

\title{Diverse Distributions of Self-Supervised Tasks\\ for Meta-Learning in NLP}

\author{Trapit Bansal$^\spadesuit$\thanks{\;\;Correspondence: tbansal@cs.umass.edu.}~~\thanks{\;\;Part of the work was done at Microsoft Research.} \and
Karthick Gunasekaran$^\spadesuit$ \and
Tong Wang$^\vardiamond$ \and \\ 
\textbf{Tsendsuren Munkhdalai}$^\clubsuit$\footnotemark[2] \and
\textbf{Andrew McCallum}$^\spadesuit$ \\
$^\spadesuit$\ University of Massachusetts Amherst\\
$^\vardiamond$\ Microsoft Research, Montr\'eal, Canada\\
$^\clubsuit$\ Google Research
}

\begin{document}
\maketitle
\begin{abstract}
Meta-learning considers the problem of learning an efficient learning process that can leverage its past experience to accurately solve new tasks. However, the efficacy of meta-learning crucially depends on the distribution of tasks available for training, and this is often assumed to be known a priori or constructed from limited supervised datasets. In this work, we aim to provide task distributions for meta-learning by considering self-supervised tasks automatically proposed from unlabeled text, to enable large-scale meta-learning in NLP. We design multiple distributions of self-supervised tasks by considering important aspects of task diversity, difficulty, type, domain, and curriculum, and investigate how they affect meta-learning performance. Our analysis shows that all these factors meaningfully alter the task distribution, some inducing significant improvements in downstream few-shot accuracy of the meta-learned models. Empirically, results on 20 downstream tasks show significant improvements in few-shot learning -- adding up to $+4.2\%$ absolute accuracy (on average) to the previous unsupervised meta-learning method, and perform comparably to supervised methods on the FewRel 2.0 benchmark.

\end{abstract}

\section{Introduction}

Humans show a remarkable capability to accurately solve a wide range of problems efficiently -- utilizing a limited amount of computation and experience. 
Deep learning models, by stark contrast, can be trained to be highly accurate on a narrow task while being highly inefficient in terms of the amount of compute and data required to reach that accuracy.
Within natural language processing (NLP), recent breakthroughs in unsupervised pretraining have enabled reusable models that can be applied to many NLP tasks, however, learning of new tasks is still inefficient \cite{yogatama2018learning,bansal2020learning,linzen2020can}.
Meta-learning \cite{schmidhuber1987evolutionary,bengio1992optimization,thrun2012learning} treats the learning process itself as a learning problem from data, with the goal of learning systems that can generalize to new tasks efficiently.
This has the potential to produce few-shot learners that can accurately solve a wide range of new tasks.
However, meta-learning requires a distribution over tasks with relevant labeled data that can be difficult to obtain, severely limiting the practical utility of meta-learning methods.

In the supervised setting, in particular, meta-learning task distribution is often defined by sub-sampling from the classes in a classification problem over a fixed dataset \cite{vinyals2016matching}.
This not only limits the applicability of meta-learning to the underlying classification problem,
but also requires a diverse set of supervised datasets with a large number of classes to enable learning.
Self-supervised meta-learning, on the other hand, seeks to propose tasks from unlabelled data \cite{hsu2018unsupervised,bansal2020self},
and has great potential to enable numerous important applications \cite{hospedales2020meta} such as neural architecture search, continual learning, hyper-parameter optimization, learning in low-resource settings, etc.
Existing work in meta-learning for NLP, however, defaults to task distributions that tend to be overly simplistic, e.g. using existing supervised datasets \cite{han2018fewrel,dou2019investigating,bansal2020learning} or unsupervised cloze-style tasks with uniform selection of words from the vocabulary \cite{bansal2020self}.
Given the lack of exploration on this critical component, we propose to devise and evaluate various task distributions in the context of unsupervised meta-learning for NLP.

Specifically, we explore a diverse set of approaches to create task distributions that are inductive to better meta-training efficacy.
We provide empirical evidence that existing definitions of task distributions are prone to producing tasks that might not be challenging enough for the underlying model to learn useful representations, which in turn translates into poor downstream task performance.
We therefore propose several new approaches that instead consider important features of the task distribution including task diversity, difficulty, resemblance to the downstream tasks, and the curriculum or the order in which tasks are presented during training.
When evaluated on a suite of 20 NLP classification tasks, our best unsupervised meta-learning method leads to an absolute increase of up to $+4.2\%$ in average few-shot accuracy over \textit{unsupervised} baseline results; and it even outperforms  \textit{supervised} meta-learning methods on FewRel 2.0 benchmark \cite{gao-etal-2019-fewrel} on 5-shot evaluation.

The paper is organized as follows.
We start by providing some relevant background (\ref{sec:background}) on meta-learning and the unsupervised task generation approach in SMLMT.
Next, we introduce (\ref{sec:methods}) new approaches to improve the task distribution. 
We then analyze (\ref{sec:analysis}) the different unsupervised distributions and how they relate to each other.
Finally, we evaluate (\ref{sec:downstream}, \ref{sec:fewrel}) the different unsupervised methods on a wide range of NLP tasks including sentiment classification, entity typing, text classification, sentence-pair classification and relation classification.


\section{Background}
\label{sec:background}
\subsection{Meta-Learning}
In this work, we focus on Model Agnostic Meta-Learning (MAML) \cite{Finn:2017:MMF:3305381.3305498}, which is an optimization-based meta-learning method.
To efficiently adapt to a task training data, MAML jointly optimizes the initial point of a neural network model and a gradient-descent based optimizer.
This is framed as a bi-level optimization consisting of an inner loop for task-specific learning and outer loop for fast adaptation across tasks:
\begin{align*}
    &\mbox{Inner:  } \theta'_i \leftarrow \theta - \alpha \nabla_{\theta} \mathcal{L}_i(\mathcal{D}^{tr}, \theta) \\
    &\mbox{Outer:  }  \Theta \leftarrow \Theta - \beta \; \nabla_{\Theta} \mathbb{E}_{T_i \sim \mathcal{P}(\mathcal{T})} \left[ L_i(\mathcal{D}^{val}, \theta'_i) \right] &
\end{align*}
where $\theta$ are the parameters of the model, $\alpha$ is the (learnable) inner loop learning rate, $\Theta := \{\theta, \alpha\}$, $\mathcal{L}_i$ is the loss for task $T_i$, $(\mathcal{D}^{tr}, \mathcal{D}^{val}) \sim T_i$ are support and validation data for the task $T_i$, and $\beta$ is the outer loop learning rate which is a hyper-parameter.
Typically, multiple steps of gradient descent are performed in the inner loop.
Training such methods proceeds in an episodic framework \cite{vinyals2016matching}, where in each episode a mini-batch of tasks are sampled along with their support and validation sets, and the model parameters are optimized as above.

\subsection{Task Distribution for Meta-Learning}
Meta-learning assumes access to a distribution $\mathcal{P(T)}$ over tasks.
The goal is to utilize tasks $T_{i} \sim \mathcal{P(T)}$ sampled from this distribution to train a learning procedure that generalizes to unseen tasks $T'\sim \mathcal{P(T)}$ from the distribution.
Supervised meta-learning often utilizes a fixed task dataset to create $\mathcal{P(T)}$ by sub-sampling from all class labels \cite{vinyals2016matching}.
\citet{bansal2020self} sought to provide an unsupervised approach that proposes tasks from unlabelled data.
The resulting Subset Masked Language Modeling Tasks (SMLMT) approach proposes self-supervised tasks to enable meta-learning and improves few-shot learning across a diverse set of classification tasks.

Sampling an $N$-way task from SMLMT requires first sampling a size-$N$ subset of the vocabulary, which are subsequently mapped to consecutive integer ids and serve as labels for the task.
Then to sample examples for each label, sentences containing that word are sampled and the occurrences of the word are masked out.
Note that a task in SMLMT is a sentence classification task where each input sentence consists of exactly one word type that is masked throughout the sentence and the label for the sentence is the underlying word type that was masked.
This enables sampling combinatorially many classification tasks for meta-learning.

\section{Exploring Unsupervised Task Distribution for Meta-Learning}
\label{sec:methods}
Sampling tasks in SMLMT depends on sampling of words, which serve as labels, and sampling of sentences containing that word.
The original formulation used uniform sampling for both steps.
This can lead to several limitations on the quality of the resulting task distribution including task diversity and difficulty.
The single-sentence classification tasks also lack cross-sentence reasoning capacities, leading to a severe train-test mismatch for downstream tasks involving sentence pairs.
To remedy these problems, we consider alternative distributions that are inductive to more diverse and challenging tasks for meta-training.
We also describe an automatic curriculum over tasks that seeks to continuously find challenging tasks for the model during training.

\subsection{Sampling labels in SMLMT}
\label{sec:cluster}
\paragraph{Frequency-based sampling:}
Word distribution in natural language is characterized by an exponential distribution with a long tail of rare words \cite{baayen2002word}.
Uniform sampling of words in SMLMT puts a disproportionately high weight on the long tail, leading to inefficient use of the training corpora since the low frequency words occur in only a small proportion of the sentences.
On the other hand, simple frequency-based sampling can be highly skewed towards a handful of high frequency words.
We thus propose to simply
sample words in proportion to their log-frequency instead.

\paragraph{Cluster-based sampling:}
Given two words randomly sampled from a large vocabulary, it is likely to be rather trivial to distinguish their corresponding contexts.
This can lead to overly simple tasks in the SMLMT task distribution.
To avoid this problem, we consider clustering words based on pre-trained word embeddings and grouping words into semantically-related clusters. 
Diverse and difficult instances of tasks in SMLMT can then be sampled by selecting all words in a task from either (1) the same cluster  (\textit{intra-cluster} sampling), or (2) different clusters (\textit{inter-cluster} sampling).
Words co-occurring in the same cluster are semantically or topically related and hence occur in similar contexts, leading to harder to classify sentences as we see in our analysis (Sec~\ref{sec:analysis}).
Moreover, choosing different clusters to sample words across tasks provides a natural diversity over topics in the training tasks.
On the other hand, picking words from different clusters (\textit{inter-cluster} sampling) can still lead to 
tasks where the sentences are easy to classify due to easily distinguishable contexts.

Specifically, clustering of pre-trained word embeddings using $k$-means has been proven effective in generating topical clusters rivaling topic models \cite{sia2020tired}.
We use the FastText \cite{joulin2017bag} embeddings as word representations. We choose FastText as it is fast, incorporates sub-word information, can generate embeddings for out-of-vocabulary words, and has been found to yield topical clusters \cite{sia2020tired}.

Since cluster sizes can be imbalanced, we pick clusters proportional to the number of words in the cluster.
Thus, assuming $\{C_1, \ldots, C_m\}$ to be the $m$ clusters of the word vocabulary, we replace the uniform sampling over words in SMLMT as:
\begin{align*}
    p_i &= \frac{|C_i|}{\sum_{t=1}^m |C_t|} \\
    i &\sim \text{Cat}(p_1, \ldots, p_m) \\
    w | i &\sim \text{Uniform}(\{w|w \in C_i\})
\end{align*}
where $\text{Cat}(p_1, \ldots, p_m)$ is a categorical distribution over $m$ categories with probabilities $\{p_1, \ldots, p_m\}$.

\subsection{Dynamic Curriculum over Self-Supervised Tasks}
\label{sec:dynamic}
The methods discussed so far use a static task distribution for learning with tasks sampled i.i.d from this distribution for training.
Curriculum learning \cite{bengio2009curriculum,graves2017automated} instead posits that choosing the order in which instances are presented, with gradually increasing complexity, can enable faster learning and better generalization.
We explore whether a curriculum in task sampling is beneficial for meta-learning by proposing a method to sample increasingly difficult tasks during training.
To enable this we need a method to propose difficult tasks based on the current state of the model during training.

Since words act as labels in SMLMT, words that are closer in the representational space of the neural model will be more difficult to distinguish, leading to more difficult tasks.
On the other hand, nearest-neighbors can be too difficult to induce effective learning for a model.
This is related to findings in negative sampling in metric learning literature \cite{schroff2015facenet,suh2019stochastic} where using “too hard” negatives typically hurts performance.

To alleviate this problem, we cluster representations computed from the model and uniformly sample words within the same cluster to create difficult but not impossible tasks (similar to the "static" clustering approach).
Secondly, we adopt an easy-to-hard curriculum by controlling the ratio between the harder tasks from the dynamic distribution $\mathcal{D}_t$ and the easier ones from the static distribution $\mathcal{S}$, consisting of tasks sampled i.i.d from uniform random word sampling or fixed word-clustering.
At step $t$, let $\lambda_t$ be the probability of sampling a task from $\mathcal{D}_t$ and $1-\lambda_t$ from $\mathcal{S}$. Then the dynamic curriculum is defined by sampling tasks from the following mixture distribution with $\lambda_t$ linearly annealed over the training epochs from $0$ to $1$:
\begin{equation*}
    T \sim \lambda_t \mathcal{D}_t + (1-\lambda_t) \mathcal{S}
\end{equation*}

To construct $\mathcal{D}_t$, we consider the following word (i.e.~label) representation for clustering, obtained by the average representation under the model of the masked sentences corresponding to a word:
\begin{equation*}
    \hat{w}^{(t)}_i = \mathbb{E}_{x\sim S(w_i)} \left[f_{\theta_t}(x) \right]
\end{equation*}
where $S(w_i)$ is the set of all sentences containing the word $w_i$ with the word $w_i$ masked out (as defined in SMLMT), $f_{\theta_t}(.)$ is the representation from the neural model for instance $x$ that is fed into the softmax classification layer, and $\theta_t$ are the model parameters at step $t$.

To make the computation of $\hat{w}^{(t)}_i$ tractable,
we first approximate the quantity by the expectation over a subset of $S(w_i)$.
Moreover, since computing the representations $\{\hat{w}^{(t)}_i\}$ for all vocabulary words and clustering at every step $t$ of the training will be computationally infeasible, we consider doing this after $m$ steps of meta-training. This also allows the model to train on the current distribution for sometime before abruptly changing the distribution.
Finally, while the model is being updated between time step $t$ and $t+m$, we use the model snapshot at $t$ to create the word clusters asynchronously for the model at $t+m$, which allows the task generation to run in parallel to the model training.

\subsection{Task proposal using sentence clustering}
\label{sec:sent_cluster}
SMLMT uses a data-augmentation strategy to automatically assign labels to unlabelled sentences by consistently masking out the \textit{same} word type in a set of sentences. The masked out word then serves as the label for the sentences.
This cloze-style approach to creating tasks was inspired by the success of masked language modeling \cite{devlin2018bert} in learning useful representations.
While this leads to significant improvements in sentence-level classification on a range of real downstream tasks \cite{bansal2020self},
it is unclear whether a word masking approach is the most efficient to learning useful sentence representations.
To probe this question further, we explore an alternative to SMLMT that directly assigns semantic labels to sentences without any augmentation.

Specifically, we consider pre-trained sentence representations for proposing tasks, which have been proven useful for improving semi-supervised learning \cite{du2020self}.
We use a pre-trained sentence embedding model \cite{du2020self,wenzek2020ccnet} to embed all sentences in a corpus and cluster them.
To propose an $N$-way task, we first randomly sample $N$ cluster-ids and remap them to random consecutive integers $\{1, \ldots, N\}$.
Then examples for each label are sampled from the corresponding cluster, creating a classification task for classifying the sentences into their underlying cluster labels.
Note that the step of remapping the cluster-ids ensures that the model cannot memorize the sentence to cluster mapping, which would lead to meta over-fitting \cite{hsu2018unsupervised}.

\subsection{Contrastive learning over sentence pairs}
\label{sec:sentpair}
SMLMT proposes sentence-level tasks and thus lacks cross-sentence reasoning.
This is confirmed by the poor downstream few-shot performance of models trained on SMLMT (see Sec.~\ref{sec:downstream}).
Since models trained on SMLMT have never seen pairs of sentences as input, it leads to a train-test mismatch for sentence-pair classification tasks.
To remedy this, we introduce a simple but effective contrastive learning task over sentence-pairs that bridges this gap.
Contrastive learning has been used to learn effective sentence representations \cite{logeswaran2018efficient}.
Next sentence prediction, a sentence-pair task, was used in the training of BERT \cite{devlin2018bert} which was later found to be not effective \cite{liu2019roberta}.
BERT considered segments instead of full sentences, however the downstream tasks often require reasoning over complete sentences.
Thus, we consider classifying whether two sentences come from the same document as opposed to different documents, as a sentence-pair task to enable cross-sentence reasoning.
This simple objective was found to be quite effective in our experiments.
Note that during meta-training, this can be treated as an additional task in the task distribution. Since the SMLMT task distribution consists of an exponential number of tasks, we sample the sentence-pair task in an episode with a fixed probability $\alpha$, which is hyper-parameter.

\section{Experiments}
We evaluate various self-supervised task distributions for their utility in meta-learning for few-shot classification.
We first describe the experimental setting,
then we perform evaluations to understand how the different self-supervised tasks relate to each other, and finally show performance on a large set of 20 real classification datasets. These datasets cover a wide range of tasks: sentiment classification, entity typing, text classification, sentence pair classification and relation classification. Our proposed approach shows significant improvements over previous few-shot classification results \cite{bansal2020self, gao-etal-2019-fewrel}.

\subsection{Experimental Setup}
We consider the challenging few-shot setting where models are trained on unlabelled corpora and then evaluated on target tasks with only $k$ examples per label ($k \le 32$) to allow fine-tuning of the models on the target task.
Since our focus is on unsupervised meta-learning, we closely follow the experimental setup of \citet{bansal2020self}.

\paragraph{Meta-learning Model}
We use the same model as in \citet{bansal2020self} for our results to be comparable\footnote{\label{code}Code and datasets will be available soon}.
The model is a BERT transformer encoder coupled with a parameter generator, a 2-layer MLP, that generates the initial point for classification layer for a task conditioned on the support examples.
The model is meta-trained using the MAML algorithm \cite{Finn:2017:MMF:3305381.3305498}, with learned per-layer learning rates, on the self-supervised task distributions.
All model hyper-parameters are kept the same so that any change in performance can be attributed to differences in the task distribution. See Supplementary for all hyper-parameters.

\paragraph{Methods Evaluated}
We consider all the different approaches to self-supervised task distributions described in Sec~\ref{sec:methods} and the baseline approach of SMLMT:
(1) \textit{Uniform}: this is the SMLMT approach of \citet{bansal2020self} which use uniform random sampling over word-types;
(2) \textit{Frequency}: SMLMT with a sampling proportional to log-frequency (see \ref{sec:cluster});
(3) \textit{Cluster}: SMLMT where labels are picked from same word cluster (see \ref{sec:cluster});
(4) \textit{Dynamic}: curriculum-based task sampling with Cluster as the static distribution (see \ref{sec:dynamic});
(5) \textit{Cluster-ccnet}: same as Cluster but using ccnet \cite{wenzek2020ccnet} as the corpora, which consists of web crawled data;
(6) \textit{SentCluster}: alternative to SMLMT which proposes tasks from subsets of sentence clustering (see \ref{sec:sent_cluster});
(7) \textit{SentPair}: the sentence-pair tasks (see \ref{sec:sentpair}).
All methods, except SentCluster and Cluster-ccnet, have Wikipedia as the text corpora. The sentence embeddings for SentCluster task distribution were obtained from \citet{du2020self}, and consist of embeddings of about 1 billion sentences from ccnet \cite{wenzek2020ccnet}. For this reason, we also report Cluster-ccnet that uses this same set of sentences.
We found it beneficial to include 25\% \textit{Frequency} tasks in the \textit{Cluster} task distribution and \textit{SentPair} tasks are included in all other task distributions unless otherwise noted.
Note that we only consider completely unsupervised meta-learning methods for fair evaluation. 
However our results improve over \citet{bansal2020self} which showed improvements over BERT and multi-task BERT baselines. As we utilize the same dataset splits released in their work, our results can be directly compared.

\subsection{Analyzing task distributions}
\label{sec:analysis}
We start by a quantitative exploration of the various self-supervised task proposals  without resorting to full fine-tuning on downstream tasks.
Our goal here is to understand properties of these task distributions and how they relate to each other.
To do this, we consider models meta-trained on a specific type of task proposal (rows in Table~\ref{tab:analysis}) and evaluate their performance in the few-shot setting on tasks sampled from all of the other task proposal methods (columns therein). We use r$i$ (or c$j$) below to refer to row $i$ (or column $j$) in the table.

We consider the following task proposal methods: Frequency (FREQ, c1): using the frequency-based word sampling in SMLMT; Inter-Cluster (X-C, c2): using the word-clustering approach explained in sec~\ref{sec:cluster} but sampling all labels of task from different clusters; Intra-Cluster (I-C, c3\&4): using the word-clustering approach explained in sec~\ref{sec:cluster} which samples all labels of task from the same cluster; Sentence Cluster (S-C, c5): this is the sentence clustering approach to task proposal presented in sec~\ref{sec:sent_cluster}.
For evaluation, we consider 4-way tasks sampled from the above methods and evaluate average accuracy over 5000 tasks.
We consider a BERT model (r1) which is not trained on the SMLMT distribution but is trained on the related masked language modeling (MLM) task. To enable evaluation of this model, we use it as a prototypical network model \cite{snell2017prototypical}.
We also consider meta-trained models trained on the SMLMT distribution with uniform sampling \cite{bansal2020self} (r2), frequency-based sampling (r3), and intra-cluster sampling (r4).
Note that all models are trained on Wikipedia corpus.

\begin{table}[t!]
\centering \fontsize{8.0}{9.5}\selectfont \setlength{\tabcolsep}{0.5em}
\begin{tabular}{lccccc}
\multirow{2}{*}{Model} & FREQ & X-C & I-C & I-C & S-C \\ 
 &  &  &  & (ccnet) & (ccnet) \\ \Xhline{2\arrayrulewidth}\\[-1.3\medskipamount]
BERT & 43.1 & 43.3 & 37.7 & 38.7 & 66.3  \\
SMLMT (uniform) & 96.2 & 96.5 & \textbf{78.4} & \textbf{68.5} & 91.7  \\
SMLMT (frequency) & 96.8 & 96.9 & \textbf{79.6} & \textbf{70.0} & 91.0 \\
SMLMT (clustering) & 96.9 & 97.0 & 96.9 & \textbf{75.2} & 94.7 \\
Sentence Cluster & 69.2 & 71.2 & \textbf{53.0} & \textbf{45.0} & 98.9 \\
\bottomrule
\end{tabular}
\caption{Analysis of task proposals. The columns are the different task proposal methods
and rows are models trained on unsupervised task distributions.
Low accuracy on a task distribution
indicates harder to classify tasks or missing information in the training distribution (see Sec~\ref{sec:analysis} for details).}
\label{tab:analysis}
\end{table}
Results are in Table~\ref{tab:analysis}.
First, since BERT wasn't trained on any of the task distributions, we find low accuracy on all these tasks on r1,
indicating that they contain information different than what is learned from MLM.
Moreover, the highest accuracy of this model is on Sentence Cluster tasks (r1c5; random baseline is 25\%), even though the domain of this task is quite different than the training data of BERT.
Next, lets consider the vanilla SMLMT model which uses uniformly random word sampling to create the meta-learning task distribution.
Interestingly, we find that it gives high accuracy on frequency-sampled tasks (r2c1).
Similarly, accuracy is high on the inter-cluster tasks (r2c2), even though the model wasn't meta-trained directly on this distribution.
More importantly, performance drops significantly ($\approx18\%$) on the tasks sampled using the intra-cluster approach (r2c3).
This performance drops even further ($\approx10\%$; r2c4) when the tasks are sampled from a different domain (common crawl) than the training domain of the model (Wiki).
Accuracy on Sentence Cluster is also very high (r2c5), without training on this distribution.
Models trained on frequency-based sampling perform similarly (r3).
We also show the performance of a model trained on tasks sampled using the intra-cluster approach. Note that this model was trained on Wikipedia corpus, and even though it was trained on intra-cluster tasks, we still see a significant performance drop on intra-cluster tasks on a different domain (r4c4 vs r4c3).
Finally, consider models trained on the sentence clustering tasks.
These perform poorly on all of the tasks proposed by SMLMT (r5c1--4), indicating that this task distribution does not contain the same amount of information as SMLMT.

In summary, these results indicate that: (1) the intra-cluster tasks are more difficult than frequency-based sampling, and inter-cluster tasks are as easy as uniform-sampling (r2c2) (2) sentence cluster tasks are the easiest among all task proposals (c5), and training on this task distribution leads to poor performance on the SMLMT distributions (r5c1--4; but not vice versa), indicating lack of information in this distribution as compared to SMLMT.
From this analysis we expect intra-cluster task distribution to be richer as compared to the other alternatives and models meta-trained on these should improve downstream performance over the others.
As we will see in the next section, the downstream performance improvements are highly correlated with these unsupervised evaluations.

\subsection{Evaluation on diverse downstream classification tasks}
\label{sec:downstream}

\paragraph{Datasets}
We consider all 17 downstream tasks in \citet{bansal2020self} and 2 additional sentence-pair tasks.
We group performance on datasets by the type of the task:
(1) \textit{Sentiment classification}: 4 domains (Books, DVD, Kitchen, Electronics) of Amazon review binary sentiment datasets \cite{blitzer2007biographies};
(2) \textit{Rating classification}:  4 domain of 3-way classification based on ratings of reviews from the above Amazon datasets, 1 dataset on 3-way classification of tweets about sentiment towards Airlines;
(3) \textit{Entity typing}: CoNLL-2003 \cite{sang2003conll} entity mention classification into 4 coarse types, MIT-Restaurant \cite{liu2013asgard} task on classifying mentions in user queries about restaurants into 8 types; 
(4) \textit{Sentence-pair classification}: Scitail, a scientific natural language inference dataset \cite{khot2018scitail}, RTE task on textual entailment and MRPC task on paraphrase classification from the GLUE benchmark \cite{wang2018glue}.
(5) \textit{Other text classification}: multiple social-media datasets on classifying tweets into (a) 2-way: political audience, bias or mention of a disaster, (b) 9-way: classifying based on political message, (c) 13-way: classifying emotion.

\paragraph{Evaluation Protocol}
We meta-train separate models on the self-supervised task distributions, without any access to the downstream supervised tasks.
The models are then fine-tuned on the downstream task training sets which consist of $k=8,16,32$ examples per class. Note that tasks can have different number of classes.
Following \citet{bansal2020self}, we use the development set of Scitail and Amazon-Electronics to select the number of steps of fine-tuning for all models, all other hyper-parameters are kept the same as meta-training.
Since few-shot performance is sensitive to the few examples in training, each model is fine-tuned on 10 sets for each task and the average test performance is reported with standard deviation.


\begin{figure}[t!]
    \centering
    \includegraphics[width=\linewidth]{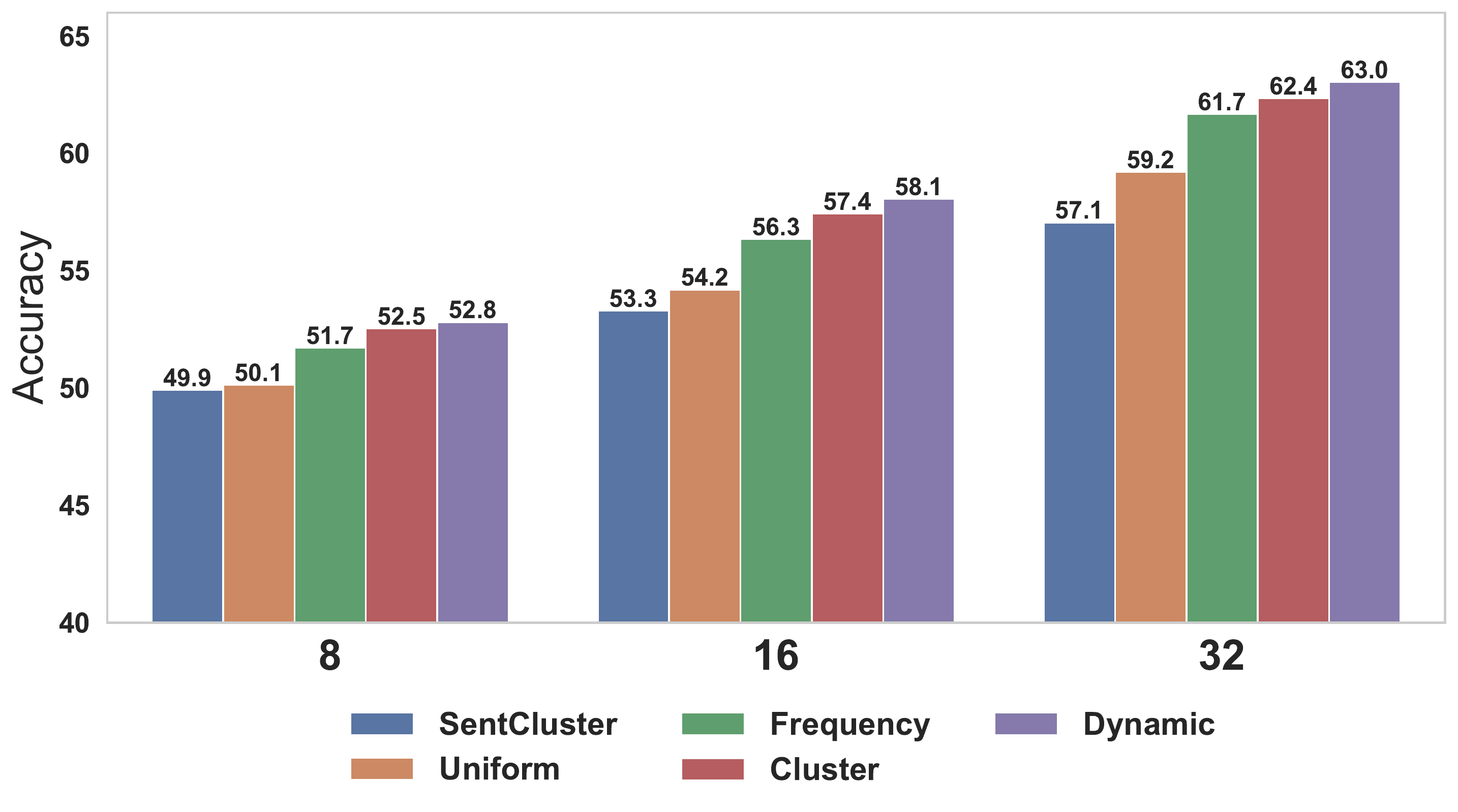}
    \caption{Overall average across 19 downstream tasks for the different task distributions proposed in this work. Cluster tasks and Dynamic curriculum lead to the best overall accuracy.}
    \label{fig:overall}
\end{figure}

\begin{figure}[t!]
    \centering
    \includegraphics[width=\linewidth]{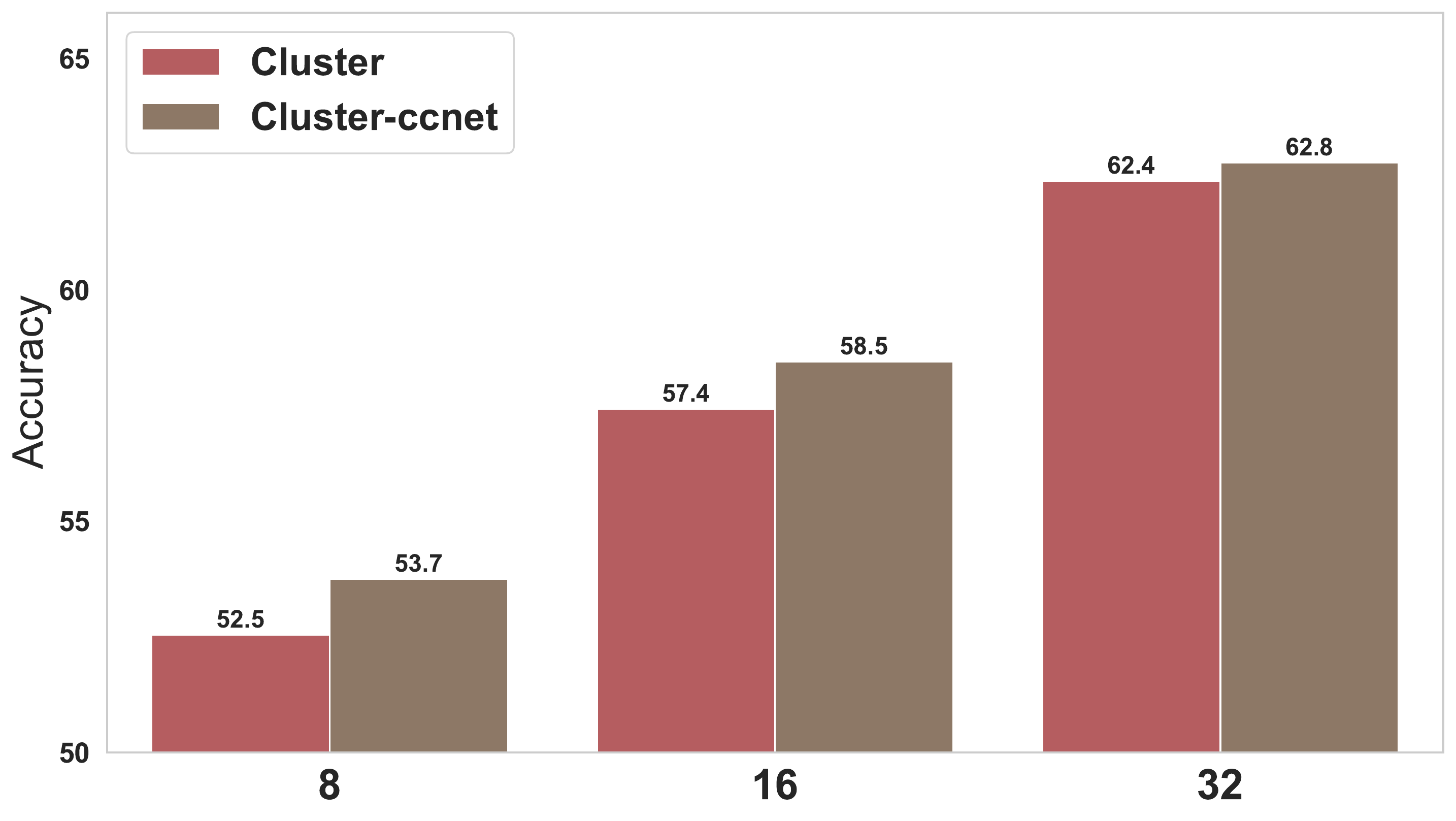}
    \caption{Changing domain of tasks from Wikipedia to CommonCrawl (ccnet) while keeping size of data, compute and model fixed. Overall average across 19 downstream tasks is shown. More diverse domain (ccnet) in training leads to improved down-stream accuracy.}
    \label{fig:domain}
\end{figure}



\begin{table}[tb!]
\centering \fontsize{8.0}{9.5}\selectfont \setlength{\tabcolsep}{0.5em}
\begin{tabular}{ccccc}
\Xhline{2\arrayrulewidth}\\[-1.3\medskipamount]
    \multirow{2}{*}{Task Group} & \multirow{2}{*}{Model} & \multicolumn{3}{c}{$k$-shot} \\ \cline{3-5}\\[-1.3\medskipamount]
      &  & $8$ & $16$ & $32$ \\[3pt] \Xhline{2\arrayrulewidth}\\[-1.3\medskipamount]
    \multirow{6}{*}{\shortstack[l]{Sentiment\\Classification}} 
       &  Uniform &
     59.1 \tiny{$\pm$ 5.2}  & 62.6 \tiny{$\pm$ 5.3} & 69.6 \tiny{$\pm$ 5.1} \\
     &  Frequency & 
    60.2 \tiny{$\pm$ 4.8} & 65.2 \tiny{$\pm$ 5.1} & 74.0 \tiny{$\pm$ 5.4} \\
     &  Cluster & 
    62.2 \tiny{$\pm$ 5.3} & 67.3 \tiny{$\pm$ 5.9} & 75.9 \tiny{$\pm$ 4.0} \\
    & Dynamic & 
     \textbf{63.6} \tiny{$\pm$ 6.0} & \textbf{69.3} \tiny{$\pm$ 6.3}  & \textbf{77.3} \tiny{$\pm$ 4.6} \\
     \cmidrule{2-2}
    & SentCluster & 
     61.1 \tiny{$\pm$ 5.8}  & 64.2 \tiny{$\pm$ 5.7} & 70.4 \tiny{$\pm$ 4.7} \\
    &  Cluster-ccnet & 
    \textbf{64.7} \tiny{$\pm$ 6.6}  & \textbf{69.9} \tiny{$\pm$ 7.1} & \textbf{76.2} \tiny{$\pm$ 6.3} \\ 
    \midrule
    \multirow{6}{*}{\shortstack[l]{Rating\\Classification}} 
       &  Uniform &
     41.9 \tiny{$\pm$ 7.2}  & 47.3 \tiny{$\pm$ 7.2} & 52.9 \tiny{$\pm$ 7.6} \\
     &  Frequency & 
     42.6 \tiny{$\pm$ 6.9} & 49.2 \tiny{$\pm$ 7.2} & 55.1 \tiny{$\pm$ 7.7}\\
     &  Cluster & 
     45.2 \tiny{$\pm$ 7.7}  & 51.9 \tiny{$\pm$ 6.6} & 56.5 \tiny{$\pm$ 7.1} \\
     & Dynamic & 
     \textbf{46.3} \tiny{$\pm$ 8.1} & \textbf{53.5} \tiny{$\pm$ 7.0}  & \textbf{57.9} \tiny{$\pm$ 7.3}  \\
     \cmidrule{2-2}
     & SentCluster & 
     45.1 \tiny{$\pm$ 8.8}  & 48.7 \tiny{$\pm$ 9.2} & 50.9 \tiny{$\pm$ 9.0} \\
     &  Cluster-ccnet & 
     \textbf{45.2} \tiny{$\pm$ 7.5}  & \textbf{52.1} \tiny{$\pm$ 7.3} & \textbf{57.1} \tiny{$\pm$ 7.8} \\ 
    \midrule
    \multirow{6}{*}{\shortstack[l]{Entity Typing}} 
       &  Uniform &
      61.4 \tiny{$\pm$ 2.6} & 72.5 \tiny{$\pm$ 4.8} & 81.4 \tiny{$\pm$ 3.0} \\
     &  Frequency & 
      64.0 \tiny{$\pm$ 3.0} & \textbf{73.0} \tiny{$\pm$ 2.2} & \textbf{82.1} \tiny{$\pm$ 2.0} \\
     &  Cluster & 
      \textbf{64.5} \tiny{$\pm$ 2.8} & 72.5 \tiny{$\pm$ 2.6} & 81.3 \tiny{$\pm$ 1.9}\\
     & Dynamic & 
      62.4 \tiny{$\pm$ 3.5} & 72.3 \tiny{$\pm$ 3.3} & 81.6 \tiny{$\pm$ 2.1}  \\
     \cmidrule{2-2}
     & SentCluster & 
      51.7 \tiny{$\pm$ 4.9} & 63.4 \tiny{$\pm$ 4.5} & 73.4 \tiny{$\pm$ 2.4} \\
     &  Cluster-ccnet & 
      \textbf{70.7} \tiny{$\pm$ 2.8} & \textbf{78.2} \tiny{$\pm$ 3.1} & \textbf{84.1} \tiny{$\pm$ 2.5} \\ 
     \midrule
         \multirow{6}{*}{\shortstack[l]{Sentence Pair\\Classification}} 
       &  Uniform &
      52.9 \tiny{$\pm$ 5.2} & 54.1 \tiny{$\pm$ 4.7} & 57.4 \tiny{$\pm$ 5.7} \\
     &  Frequency & 
      \textbf{59.5} \tiny{$\pm$ 7.2} & \textbf{61.0} \tiny{$\pm$ 8.5} & \textbf{63.6} \tiny{$\pm$ 9.1} \\
     &  Cluster & 
      56.4 \tiny{$\pm$ 5.3} & 59.5 \tiny{$\pm$ 7.6} & 62.8 \tiny{$\pm$ 8.6} \\
     & Dynamic & 
      55.0 \tiny{$\pm$ 4.9} & 57.8 \tiny{$\pm$ 5.7}  &  62.2 \tiny{$\pm$ 8.5} \\
      \cmidrule{2-2}
     & SentCluster & 
      52.6 \tiny{$\pm$ 4.7} & 52.9 \tiny{$\pm$ 2.9} & 54.0 \tiny{$\pm$ 3.8}\\
     &  Cluster-ccnet & 
      \textbf{55.9} \tiny{$\pm$ 5.7} & \textbf{58.5} \tiny{$\pm$ 6.9} & \textbf{62.9} \tiny{$\pm$ 6.9} \\ 
      \midrule
         \multirow{6}{*}{\shortstack[l]{Other Text\\Classification}} 
       &  Uniform &
       44.8 \tiny{$\pm$ 3.9} & 47.5 \tiny{$\pm$ 2.3} & 49.4 \tiny{$\pm$ 2.1} \\
     &  Frequency & 
     44.4 \tiny{$\pm$ 3.5} & 47.3 \tiny{$\pm$ 2.0} & 49.1 \tiny{$\pm$ 1.9} \\
     &  Cluster & 
     45.0 \tiny{$\pm$ 3.7} & 48.1 \tiny{$\pm$ 2.0} & 49.5 \tiny{$\pm$ 1.9} \\
     & Dynamic & 
     \textbf{45.5} \tiny{$\pm$ 3.5} & \textbf{48.5} \tiny{$\pm$ 2.2}  &  \textbf{49.8} \tiny{$\pm$ 1.9} \\
     \cmidrule{2-2}
     & SentCluster & 
     43.5 \tiny{$\pm$ 4.1} & 45.7 \tiny{$\pm$ 2.5} & 47.8 \tiny{$\pm$ 1.7} \\
     &  Cluster-ccnet & 
     \textbf{46.6} \tiny{$\pm$ 3.4} & \textbf{48.9} \tiny{$\pm$ 2.2} & \textbf{49.9} \tiny{$\pm$ 1.8} \\
     \midrule
\bottomrule
\end{tabular}
\caption{Results on downstream tasks. Best performing model for each $k$ and each task group is in bold and the second best is underlined.}
\label{tab:results}
\end{table}
\paragraph{Results.}
Table \ref{tab:results} shows the performance of all methods on different types of downstream tasks.
We group datasets based on task type as described above and report the average performance over all the datasets in each group.
First, note that the SentCluster approach is always inferior to any of the cloze-style approach, except on sentiment and rating classification where it is slightly better than SMLMT with Uniform sampling but worse than the other methods proposed here.
Interestingly, replacing Uniform sampling with the simple Frequency sampling already leads to significant improvements throughout. 
Comparing the Cluster approach, we observe that this is better than Frequency on sentence-level tasks (like sentiment, rating, others), while slightly worse or comparable on sentence-pair tasks and phrase-level classification tasks (entity typing).
Overall, the word-clustering approach to sampling labels for SMLMT are more preferable as they are often among the two highest performing on any task group or close to the highest performance.
Note that our unsupervised analysis in Sec~\ref{sec:analysis} also reflected that training on the Cluster task distribution should be better compared to others.
Finally, note that using the Dynamic curriculum over task sampling further improves the performance over cluster-based approach.
This overall trend is also clearly reflected in the overall average performance across all the 19 tasks in Figure~\ref{fig:overall}. 
Figure~\ref{fig:domain} further shows that, for the Cluster tasks, constructing tasks from more diverse domain such as CommonCrawl can improve downstream performance even when using the same amount of data for training.

\paragraph{Ablation over SentPair.}
We introduced the sentence pair task to enable better learning of sentence pair tasks such as natural language inference.
These task remove the train-test mismatch in the input format as the sentence pair tasks contain pairs of sentences as input where as SMLMT only proposes single sentence classification tasks.
To assess the efficacy of the SentPair task, we trained a word-cluster model with and without the SentPair task and evaluated it on few-shot sentence pair tasks of Scitail and MRPC.
Results are in Table \ref{tab:sentpair}. 
We can see that the unsupervised SentPair task improves performance under most settings, sometimes by large margins up to 8\% absolute.
\begin{table}[tb!]
\centering \fontsize{8.0}{9.5}\selectfont \setlength{\tabcolsep}{0.5em}
\begin{tabular}{ccccc}
\Xhline{2\arrayrulewidth}\\[-1.3\medskipamount]
    Task & Model & $8$ & $16$ & $32$ \\[3pt] \Xhline{2\arrayrulewidth}\\[-1.3\medskipamount]
    \multirow{2}{*}{\shortstack[l]{MRPC}} & Clustering & 
     54.63 \tiny{$\pm$ 4.69}  & 54.00 \tiny{$\pm$ 3.63} & 58.28 \tiny{$\pm$ 4.90} \\
       &  \quad\quad + SentPair &
     \textbf{55.88} \tiny{$\pm$ 6.68}  & \textbf{57.13} \tiny{$\pm$ 5.15} & \textbf{60.12} \tiny{$\pm$ 3.58} \\ \midrule
    \multirow{2}{*}{\shortstack[l]{Scitail}} & Clustering & 
     \textbf{60.63} \tiny{$\pm$ 4.29}  & 59.89 \tiny{$\pm$ 4.20} & 67.89 \tiny{$\pm$ 5.59} \\
       &  \quad\quad + SentPair &
     58.86 \tiny{$\pm$ 4.81}  & \textbf{67.94} \tiny{$\pm$ 2.92} & \textbf{73.56} \tiny{$\pm$ 2.79} \\ \midrule
\bottomrule
\end{tabular}
\caption{Ablation: training with \& without SentPair.}
\label{tab:sentpair}
\end{table}

\begin{table}[tb!]
\centering \fontsize{8.0}{9.5}\selectfont \setlength{\tabcolsep}{0.5em}
\begin{tabular}{ccccc}
\Xhline{2\arrayrulewidth}\\[-1.3\medskipamount]
     Static Distribution & $\lambda_t$ & $8$-shot & $16$-shot & $32$-shot \\[3pt] \Xhline{2\arrayrulewidth}\\[-1.3\medskipamount]
     Cluster & 0.5 & 54.0 & 58.7 & 63.0 \\
     Cluster & 0.25 & 55.2 & 59.0 & 64.6 \\
     Frequency & Anneal & 56.5 & 60.9 & 65.8 \\
     Cluster & Anneal & \textbf{56.8} & \textbf{61.7} & \textbf{66.4} \\
\bottomrule
\end{tabular}
\caption{Ablation: static tasks and the value of mixing proportion $\lambda_t$ used in dynamic curriculum. }
\label{tab:dynamic}
\end{table}
\paragraph{Ablation for dynamic curriculum.}
The dynamic curriculum over tasks requires two crucial choices: the static distribution and the value of mixing proportion $\lambda_t$.
We ablate over choices for these in Fig.~\ref{tab:dynamic} which reports average performance over 5 tasks, one each from the task groups considered. We find that using the Cluster tasks, created from static pre-computer word-embeddings, works better than using Frequency-based tasks as the static distribution. Moreover, annealing $\lambda_t$ from 0 to 1 over the training epochs is better than using a fixed value of $\lambda_t$ throughout training.

\subsection{Evaluation on FewRel 2.0 benchmark}
\label{sec:fewrel}
FewRel \cite{han2018fewrel,gao-etal-2019-fewrel} is a common benchmark for few-shot learning in NLP, which consists of many few-shot relation classification tasks created by sub-sampling from a pool of relation labels. The resemblance to the popular few-shot benchmarks like MiniImageNet \cite{vinyals2016matching} makes FewRel one of the few widely used datasets for training and evaluating NLP meta-learning methods.

\begin{table}[tb!]
\centering \fontsize{8.0}{9.5}\selectfont \setlength{\tabcolsep}{0.5em}
\begin{tabular}{lcccc}
\Xhline{2\arrayrulewidth}\\[-1.3\medskipamount]
    \multirow{2}{*}{Model} & \multicolumn{2}{c}{1-shot} & \multicolumn{2}{c}{5-shot} \\
     & $N=5$ & $N=10$ & $N=5$ & $N=10$\\[3pt] \Xhline{2\arrayrulewidth}\\[-1.3\medskipamount]
    \shortstack[l]{\textit{Unsupervised}} &  &  &  \\ \cline{1-1}\\[-1.3\medskipamount]
    Cluster &  60.1
       & 44.1 & 78.8 & 65.2 \\
    Cluster-ccnet &  61.3
       & 46.1  & \underline{80.4}  & \underline{67.7} \\[3pt]
    \shortstack[l]{\textit{Supervised}} &  &  &  \\ \cline{1-1}\\[-1.3\medskipamount]
    Proto-Adv (CNN) & 42.2
       & 28.9 & 58.7 & 44.4 \\
    Proto-Adv (BERT) & 41.9
       & 27.4 & 54.7 & 37.4 \\
    BERT-Pair & \underline{67.4}
       & \textbf{54.9} & 78.6 & 66.9 \\
    Cluster-ccnet &  \textbf{67.7}
       & \underline{52.9}  & \textbf{84.3}  & \textbf{74.1} \\ 
\bottomrule
\end{tabular}
\caption{Results on Fewrel 2.0 test set.}
\label{tab:fewrel_test}
\end{table}

Before submitting to the competition site for test set results, we first use the validation set to select the best model(s),
where we observed that the Cluster approaches performs better than the other task proposals (see validation results in Supplementary).
We then compare their test set performance with previously published results:
the BERT-Pair, Proto-Adversarial (CNN), and Proto-Adversarial (BERT) are \textit{supervised} meta-learning models trained on FewRel training data and using BERT or CNN as the text encoder.
See \citet{gao-etal-2019-fewrel} for details.
Interestingly, our \textit{unsupervised} meta-learned models that do not use any FewRel training data outperform the supervised baselines in the 5-shot setting. Performance is lower than BERT-Pair on 1-shot tasks, potentially because our models have not been trained for 1-shot tasks like BERT-Pair.
Finally, fine-tuning our best model on the FewRel training data leads to the best overall performance.

\section{Related Work}
Meta-learning applications in NLP have yielded improvements on specific tasks \cite{gu2018meta,chen2018meta,guo2018multi,yu2018diverse,han2018fewrel,dou2019investigating}.
Unsupervised meta-learning has been explored in computer vision \cite{hsu2018unsupervised,khodadadeh2019unsupervised} and reinforcement learning \cite{gupta2018unsupervised}. \citet{hsu2018unsupervised} cluster images using pre-trained embeddings to create tasks.
\citet{metz2018learning} meta-learn an unsupervised update rule
in a semi-supervised framework.
\citet{bansal2020self} developed the SMLMT approach to unsupervised meta-learning in NLP.
Contemporary work \cite{murty2021dreca} explored the use of clustering, though focused only on natural language inference tasks.
Curriculum learning \cite{bengio2009curriculum} in the context of meta-learning has been unexplored in NLP, prior to this work.
\citet{NEURIPS2019_d5a28f81} found unsupervised curriculum to be beneficial for meta-reinforcement learning.
We refer to \citet{hospedales2020meta} for a comprehensive review of meta-learning.

Self-supervised learning has emerged as an efficient approach to representation learning in NLP \cite{howard2018universal,peters2018deep,devlin2018bert,radford2019language,yang2019xlnet}. 
Multi-task learning of pre-trained models has shown improved results on many tasks \cite{phang2018sentence,liu2019multi}, including few-shot setting.
\citet{yin2020universal} leveraged entailment tasks for few-shot learning.
\citet{du2020self} developed self-training methods for semi-supervised few-shot learning.
Recently, extremely large language models have been shown to have few-shot capacities \cite{brown2020language}, while 
\citet{schick2020s} demonstrated few-shot capacities for small models in the semi-supervised setting. 
Meanwhile, \citet{bansal2020learning, bansal2020self} showed meta-learning to be effective at improving few-shot performance in multi-task and unsupervised settings, as well as improving performance for small models.

\section{Conclusion}
We explored several approaches to self-supervised task distribution for meta-learning.
Our results demonstrate improvements in few-shot performance over a wide-range of classification tasks.
This demonstrates the utility of meta-learning from unlabeled data,
opening up the possibility of large-scale meta-learning for pertinent applications in NLP
such as continual learning, architecture search, learning for low-resource languages, and more.

\section*{Acknowledgements}
This work was supported in part by the Chan Zuckerberg Initiative, in part by IBM Research AI through the AI Horizons Network, and in part by the National Science Foundation under Award No. 1763618. Any opinions, findings and conclusions or recommendations expressed in this material are those of the authors and do not necessarily reflect those of the sponsor.

\bibliography{references,task_references}

\begin{thebibliography}{50}
\expandafter\ifx\csname natexlab\endcsname\relax\def\natexlab#1{#1}\fi

\bibitem[{Baayen(2002)}]{baayen2002word}
R~Harald Baayen. 2002.
\newblock \emph{Word frequency distributions}, volume~18.
\newblock Springer Science \& Business Media.

\bibitem[{Bansal et~al.(2020{\natexlab{a}})Bansal, Jha, and
  McCallum}]{bansal2020learning}
Trapit Bansal, Rishikesh Jha, and Andrew McCallum. 2020{\natexlab{a}}.
\newblock Learning to few-shot learn across diverse natural language
  classification tasks.
\newblock In \emph{Proceedings of the 28th International Conference on
  Computational Linguistics}, pages 5108--5123.

\bibitem[{Bansal et~al.(2020{\natexlab{b}})Bansal, Jha, Munkhdalai, and
  McCallum}]{bansal2020self}
Trapit Bansal, Rishikesh Jha, Tsendsuren Munkhdalai, and Andrew McCallum.
  2020{\natexlab{b}}.
\newblock Self-supervised meta-learning for few-shot natural language
  classification tasks.
\newblock In \emph{Proceedings of the 2020 Conference on Empirical Methods in
  Natural Language Processing (EMNLP)}, pages 522--534.

\bibitem[{Bengio et~al.(1992)Bengio, Bengio, Cloutier, and
  Gecsei}]{bengio1992optimization}
Samy Bengio, Yoshua Bengio, Jocelyn Cloutier, and Jan Gecsei. 1992.
\newblock On the optimization of a synaptic learning rule.
\newblock In \emph{Preprints Conf. Optimality in Artificial and Biological
  Neural Networks}, pages 6--8. Univ. of Texas.

\bibitem[{Bengio et~al.(2009)Bengio, Louradour, Collobert, and
  Weston}]{bengio2009curriculum}
Yoshua Bengio, J{\'e}r{\^o}me Louradour, Ronan Collobert, and Jason Weston.
  2009.
\newblock Curriculum learning.
\newblock In \emph{Proceedings of the 26th annual international conference on
  machine learning}, pages 41--48.

\bibitem[{Blitzer et~al.(2007)Blitzer, Dredze, and
  Pereira}]{blitzer2007biographies}
John Blitzer, Mark Dredze, and Fernando Pereira. 2007.
\newblock Biographies, bollywood, boom-boxes and blenders: Domain adaptation
  for sentiment classification.
\newblock In \emph{Proceedings of the 45th annual meeting of the association of
  computational linguistics}, pages 440--447.

\bibitem[{Brown et~al.(2020)Brown, Mann, Ryder, Subbiah, Kaplan, Dhariwal,
  Neelakantan, Shyam, Sastry, Askell et~al.}]{brown2020language}
Tom~B Brown, Benjamin Mann, Nick Ryder, Melanie Subbiah, Jared Kaplan, Prafulla
  Dhariwal, Arvind Neelakantan, Pranav Shyam, Girish Sastry, Amanda Askell,
  et~al. 2020.
\newblock Language models are few-shot learners.
\newblock \emph{arXiv preprint arXiv:2005.14165}.

\bibitem[{Chen et~al.(2018)Chen, Qiu, Liu, and Huang}]{chen2018meta}
Junkun Chen, Xipeng Qiu, Pengfei Liu, and Xuanjing Huang. 2018.
\newblock Meta multi-task learning for sequence modeling.
\newblock In \emph{Thirty-Second AAAI Conference on Artificial Intelligence}.

\bibitem[{Devlin et~al.(2018)Devlin, Chang, Lee, and
  Toutanova}]{devlin2018bert}
Jacob Devlin, Ming-Wei Chang, Kenton Lee, and Kristina Toutanova. 2018.
\newblock Bert: Pre-training of deep bidirectional transformers for language
  understanding.
\newblock \emph{arXiv preprint arXiv:1810.04805}.

\bibitem[{Dou et~al.(2019)Dou, Yu, and Anastasopoulos}]{dou2019investigating}
Zi-Yi Dou, Keyi Yu, and Antonios Anastasopoulos. 2019.
\newblock Investigating meta-learning algorithms for low-resource natural
  language understanding tasks.
\newblock In \emph{Proceedings of the 2019 Conference on Empirical Methods in
  Natural Language Processing and the 9th International Joint Conference on
  Natural Language Processing (EMNLP-IJCNLP)}, pages 1192--1197.

\bibitem[{Du et~al.(2020)Du, Grave, Gunel, Chaudhary, Celebi, Auli, Stoyanov,
  and Conneau}]{du2020self}
Jingfei Du, Edouard Grave, Beliz Gunel, Vishrav Chaudhary, Onur Celebi, Michael
  Auli, Ves Stoyanov, and Alexis Conneau. 2020.
\newblock Self-training improves pre-training for natural language
  understanding.
\newblock \emph{arXiv preprint arXiv:2010.02194}.

\bibitem[{Finn et~al.(2017)Finn, Abbeel, and
  Levine}]{Finn:2017:MMF:3305381.3305498}
Chelsea Finn, Pieter Abbeel, and Sergey Levine. 2017.
\newblock Model-agnostic meta-learning for fast adaptation of deep networks.
\newblock In \emph{International Conference on Machine Learning}, pages
  1126--1135.

\bibitem[{Gao et~al.(2019)Gao, Han, Zhu, Liu, Li, Sun, and
  Zhou}]{gao-etal-2019-fewrel}
Tianyu Gao, Xu~Han, Hao Zhu, Zhiyuan Liu, Peng Li, Maosong Sun, and Jie Zhou.
  2019.
\newblock \href {https://doi.org/10.18653/v1/D19-1649} {{F}ew{R}el 2.0: Towards
  more challenging few-shot relation classification}.
\newblock In \emph{Proceedings of the 2019 Conference on Empirical Methods in
  Natural Language Processing and the 9th International Joint Conference on
  Natural Language Processing (EMNLP-IJCNLP)}, pages 6251--6256, Hong Kong,
  China. Association for Computational Linguistics.

\bibitem[{Graves et~al.(2017)Graves, Bellemare, Menick, Munos, and
  Kavukcuoglu}]{graves2017automated}
Alex Graves, Marc~G Bellemare, Jacob Menick, Remi Munos, and Koray Kavukcuoglu.
  2017.
\newblock Automated curriculum learning for neural networks.
\newblock In \emph{International Conference on Machine Learning}, pages
  1311--1320. PMLR.

\bibitem[{Gu et~al.(2018)Gu, Wang, Chen, Cho, and Li}]{gu2018meta}
Jiatao Gu, Yong Wang, Yun Chen, Kyunghyun Cho, and Victor~OK Li. 2018.
\newblock Meta-learning for low-resource neural machine translation.
\newblock \emph{arXiv preprint arXiv:1808.08437}.

\bibitem[{Guo et~al.(2018)Guo, Shah, and Barzilay}]{guo2018multi}
Jiang Guo, Darsh Shah, and Regina Barzilay. 2018.
\newblock Multi-source domain adaptation with mixture of experts.
\newblock In \emph{Proceedings of the 2018 Conference on Empirical Methods in
  Natural Language Processing}, pages 4694--4703.

\bibitem[{Gupta et~al.(2018)Gupta, Eysenbach, Finn, and
  Levine}]{gupta2018unsupervised}
Abhishek Gupta, Benjamin Eysenbach, Chelsea Finn, and Sergey Levine. 2018.
\newblock Unsupervised meta-learning for reinforcement learning.
\newblock \emph{arXiv preprint arXiv:1806.04640}.

\bibitem[{Han et~al.(2018)Han, Zhu, Yu, Wang, Yao, Liu, and
  Sun}]{han2018fewrel}
Xu~Han, Hao Zhu, Pengfei Yu, Ziyun Wang, Yuan Yao, Zhiyuan Liu, and Maosong
  Sun. 2018.
\newblock Fewrel: A large-scale supervised few-shot relation classification
  dataset with state-of-the-art evaluation.
\newblock In \emph{Empirical Methods in Natural Language Processing}, pages
  4803--4809.

\bibitem[{Hospedales et~al.(2020)Hospedales, Antoniou, Micaelli, and
  Storkey}]{hospedales2020meta}
Timothy Hospedales, Antreas Antoniou, Paul Micaelli, and Amos Storkey. 2020.
\newblock Meta-learning in neural networks: A survey.
\newblock \emph{arXiv preprint arXiv:2004.05439}.

\bibitem[{Howard and Ruder(2018)}]{howard2018universal}
Jeremy Howard and Sebastian Ruder. 2018.
\newblock Universal language model fine-tuning for text classification.
\newblock In \emph{Proceedings of the 56th Annual Meeting of the Association
  for Computational Linguistics (Volume 1: Long Papers)}, pages 328--339.

\bibitem[{Hsu et~al.(2019)Hsu, Levine, and Finn}]{hsu2018unsupervised}
Kyle Hsu, Sergey Levine, and Chelsea Finn. 2019.
\newblock \href {https://openreview.net/forum?id=r1My6sR9tX} {Unsupervised
  learning via meta-learning}.
\newblock In \emph{International Conference on Learning Representations}.

\bibitem[{Jabri et~al.(2019)Jabri, Hsu, Gupta, Eysenbach, Levine, and
  Finn}]{NEURIPS2019_d5a28f81}
Allan Jabri, Kyle Hsu, Abhishek Gupta, Ben Eysenbach, Sergey Levine, and
  Chelsea Finn. 2019.
\newblock \href
  {https://proceedings.neurips.cc/paper/2019/file/d5a28f81834b6df2b6db6d3e5e2635c7-Paper.pdf}
  {Unsupervised curricula for visual meta-reinforcement learning}.
\newblock In \emph{Advances in Neural Information Processing Systems},
  volume~32, pages 10519--10531. Curran Associates, Inc.

\bibitem[{Joulin et~al.(2017)Joulin, Grave, Bojanowski, and
  Mikolov}]{joulin2017bag}
Armand Joulin, {\'E}douard Grave, Piotr Bojanowski, and Tom{\'a}{\v{s}}
  Mikolov. 2017.
\newblock Bag of tricks for efficient text classification.
\newblock In \emph{Proceedings of the 15th Conference of the European Chapter
  of the Association for Computational Linguistics: Volume 2, Short Papers},
  pages 427--431.

\bibitem[{Khodadadeh et~al.(2019)Khodadadeh, Boloni, and
  Shah}]{khodadadeh2019unsupervised}
Siavash Khodadadeh, Ladislau Boloni, and Mubarak Shah. 2019.
\newblock Unsupervised meta-learning for few-shot image classification.
\newblock \emph{Advances in neural information processing systems}, 32.

\bibitem[{Khot et~al.(2018)Khot, Sabharwal, and Clark}]{khot2018scitail}
Tushar Khot, Ashish Sabharwal, and Peter Clark. 2018.
\newblock Scitail: A textual entailment dataset from science question
  answering.
\newblock In \emph{Thirty-Second AAAI Conference on Artificial Intelligence}.

\bibitem[{Linzen(2020)}]{linzen2020can}
Tal Linzen. 2020.
\newblock How can we accelerate progress towards human-like linguistic
  generalization?
\newblock In \emph{Proceedings of the 58th Annual Meeting of the Association
  for Computational Linguistics}, pages 5210--5217.

\bibitem[{Liu et~al.(2013)Liu, Pasupat, Cyphers, and Glass}]{liu2013asgard}
Jingjing Liu, Panupong Pasupat, Scott Cyphers, and Jim Glass. 2013.
\newblock Asgard: A portable architecture for multilingual dialogue systems.
\newblock In \emph{2013 IEEE International Conference on Acoustics, Speech and
  Signal Processing}, pages 8386--8390. IEEE.

\bibitem[{Liu et~al.(2019{\natexlab{a}})Liu, He, Chen, and Gao}]{liu2019multi}
Xiaodong Liu, Pengcheng He, Weizhu Chen, and Jianfeng Gao. 2019{\natexlab{a}}.
\newblock Multi-task deep neural networks for natural language understanding.
\newblock \emph{arXiv preprint arXiv:1901.11504}.

\bibitem[{Liu et~al.(2019{\natexlab{b}})Liu, Ott, Goyal, Du, Joshi, Chen, Levy,
  Lewis, Zettlemoyer, and Stoyanov}]{liu2019roberta}
Yinhan Liu, Myle Ott, Naman Goyal, Jingfei Du, Mandar Joshi, Danqi Chen, Omer
  Levy, Mike Lewis, Luke Zettlemoyer, and Veselin Stoyanov. 2019{\natexlab{b}}.
\newblock Roberta: A robustly optimized bert pretraining approach.
\newblock \emph{arXiv preprint arXiv:1907.11692}.

\bibitem[{Logeswaran and Lee(2018)}]{logeswaran2018efficient}
Lajanugen Logeswaran and Honglak Lee. 2018.
\newblock An efficient framework for learning sentence representations.
\newblock In \emph{International Conference on Learning Representations}.

\bibitem[{Metz et~al.(2019)Metz, Maheswaranathan, Cheung, and
  Sohl-Dickstein}]{metz2018learning}
Luke Metz, Niru Maheswaranathan, Brian Cheung, and Jascha Sohl-Dickstein. 2019.
\newblock \href {https://openreview.net/forum?id=HkNDsiC9KQ} {Learning
  unsupervised learning rules}.
\newblock In \emph{International Conference on Learning Representations}.

\bibitem[{Murty et~al.(2021)Murty, Hashimoto, and Manning}]{murty2021dreca}
Shikhar Murty, Tatsunori Hashimoto, and Christopher~D Manning. 2021.
\newblock Dreca: A general task augmentation strategy for few-shot natural
  language inference.
\newblock In \emph{Proceedings of the 2021 Conference of the North American
  Chapter of the Association for Computational Linguistics: Human Language
  Technologies}, pages 1113--1125.

\bibitem[{Peters et~al.(2018)Peters, Neumann, Iyyer, Gardner, Clark, Lee, and
  Zettlemoyer}]{peters2018deep}
Matthew~E Peters, Mark Neumann, Mohit Iyyer, Matt Gardner, Christopher Clark,
  Kenton Lee, and Luke Zettlemoyer. 2018.
\newblock Deep contextualized word representations.
\newblock In \emph{Proceedings of NAACL-HLT}, pages 2227--2237.

\bibitem[{Phang et~al.(2018)Phang, F{\'e}vry, and Bowman}]{phang2018sentence}
Jason Phang, Thibault F{\'e}vry, and Samuel~R Bowman. 2018.
\newblock Sentence encoders on stilts: Supplementary training on intermediate
  labeled-data tasks.
\newblock \emph{arXiv preprint arXiv:1811.01088}.

\bibitem[{Radford et~al.(2019)Radford, Wu, Child, Luan, Amodei, and
  Sutskever}]{radford2019language}
Alec Radford, Jeffrey Wu, Rewon Child, David Luan, Dario Amodei, and Ilya
  Sutskever. 2019.
\newblock Language models are unsupervised multitask learners.
\newblock \emph{OpenAI Blog}, 1(8).

\bibitem[{Sang and De~Meulder(2003)}]{sang2003conll}
Erik F Tjong~Kim Sang and Fien De~Meulder. 2003.
\newblock Introduction to the conll-2003 shared task: Language-independent
  named entity recognition.
\newblock In \emph{Proceedings of the Seventh Conference on Natural Language
  Learning at HLT-NAACL 2003}, pages 142--147.

\bibitem[{Schick and Sch{\"u}tze(2020)}]{schick2020s}
Timo Schick and Hinrich Sch{\"u}tze. 2020.
\newblock It's not just size that matters: Small language models are also
  few-shot learners.
\newblock \emph{arXiv preprint arXiv:2009.07118}.

\bibitem[{Schmidhuber(1987)}]{schmidhuber1987evolutionary}
J{\"u}rgen Schmidhuber. 1987.
\newblock \emph{Evolutionary principles in self-referential learning, or on
  learning how to learn: the meta-meta-... hook}.
\newblock Ph.D. thesis, Technische Universit{\"a}t M{\"u}nchen.

\bibitem[{Schroff et~al.(2015)Schroff, Kalenichenko, and
  Philbin}]{schroff2015facenet}
Florian Schroff, Dmitry Kalenichenko, and James Philbin. 2015.
\newblock Facenet: A unified embedding for face recognition and clustering.
\newblock In \emph{Proceedings of the IEEE conference on computer vision and
  pattern recognition}, pages 815--823.

\bibitem[{Sia et~al.(2020)Sia, Dalmia, and Mielke}]{sia2020tired}
Suzanna Sia, Ayush Dalmia, and Sabrina~J Mielke. 2020.
\newblock Tired of topic models? clusters of pretrained word embeddings make
  for fast and good topics too!
\newblock In \emph{Proceedings of the 2020 Conference on Empirical Methods in
  Natural Language Processing (EMNLP)}, pages 1728--1736.

\bibitem[{Snell et~al.(2017)Snell, Swersky, and Zemel}]{snell2017prototypical}
Jake Snell, Kevin Swersky, and Richard Zemel. 2017.
\newblock Prototypical networks for few-shot learning.
\newblock In \emph{Advances in Neural Information Processing Systems}, pages
  4077--4087.

\bibitem[{Suh et~al.(2019)Suh, Han, Kim, and Lee}]{suh2019stochastic}
Yumin Suh, Bohyung Han, Wonsik Kim, and Kyoung~Mu Lee. 2019.
\newblock Stochastic class-based hard example mining for deep metric learning.
\newblock In \emph{Proceedings of the IEEE/CVF Conference on Computer Vision
  and Pattern Recognition}, pages 7251--7259.

\bibitem[{Thrun and Pratt(2012)}]{thrun2012learning}
Sebastian Thrun and Lorien Pratt. 2012.
\newblock \emph{Learning to learn}.
\newblock Springer Science \& Business Media.

\bibitem[{Vinyals et~al.(2016)Vinyals, Blundell, Lillicrap, Wierstra
  et~al.}]{vinyals2016matching}
Oriol Vinyals, Charles Blundell, Timothy Lillicrap, Daan Wierstra, et~al. 2016.
\newblock Matching networks for one shot learning.
\newblock In \emph{Advances in neural information processing systems}, pages
  3630--3638.

\bibitem[{Wang et~al.(2018)Wang, Singh, Michael, Hill, Levy, and
  Bowman}]{wang2018glue}
Alex Wang, Amanpreet Singh, Julian Michael, Felix Hill, Omer Levy, and Samuel~R
  Bowman. 2018.
\newblock Glue: A multi-task benchmark and analysis platform for natural
  language understanding.
\newblock \emph{arXiv preprint arXiv:1804.07461}.

\bibitem[{Wenzek et~al.(2020)Wenzek, Lachaux, Conneau, Chaudhary, Guzm{\'a}n,
  Joulin, and Grave}]{wenzek2020ccnet}
Guillaume Wenzek, Marie-Anne Lachaux, Alexis Conneau, Vishrav Chaudhary,
  Francisco Guzm{\'a}n, Armand Joulin, and {\'E}douard Grave. 2020.
\newblock Ccnet: Extracting high quality monolingual datasets from web crawl
  data.
\newblock In \emph{Proceedings of The 12th Language Resources and Evaluation
  Conference}, pages 4003--4012.

\bibitem[{Yang et~al.(2019)Yang, Dai, Yang, Carbonell, Salakhutdinov, and
  Le}]{yang2019xlnet}
Zhilin Yang, Zihang Dai, Yiming Yang, Jaime Carbonell, Russ~R Salakhutdinov,
  and Quoc~V Le. 2019.
\newblock Xlnet: Generalized autoregressive pretraining for language
  understanding.
\newblock In \emph{Advances in neural information processing systems}, pages
  5754--5764.

\bibitem[{Yin et~al.(2020)Yin, Rajani, Radev, Socher, and
  Xiong}]{yin2020universal}
Wenpeng Yin, Nazneen~Fatema Rajani, Dragomir Radev, Richard Socher, and Caiming
  Xiong. 2020.
\newblock Universal natural language processing with limited annotations: Try
  few-shot textual entailment as a start.
\newblock In \emph{Proceedings of the 2020 Conference on Empirical Methods in
  Natural Language Processing (EMNLP)}, pages 8229--8239.

\bibitem[{Yogatama et~al.(2019)Yogatama, de~Masson~d'Autume, Connor,
  Kocisk{\'{y}}, Chrzanowski, Kong, Lazaridou, Ling, Yu, Dyer, and
  Blunsom}]{yogatama2018learning}
Dani Yogatama, Cyprien de~Masson~d'Autume, Jerome Connor, Tom{\'{a}}s
  Kocisk{\'{y}}, Mike Chrzanowski, Lingpeng Kong, Angeliki Lazaridou, Wang
  Ling, Lei Yu, Chris Dyer, and Phil Blunsom. 2019.
\newblock \href {http://arxiv.org/abs/1901.11373} {Learning and evaluating
  general linguistic intelligence}.
\newblock \emph{CoRR}, abs/1901.11373.

\bibitem[{Yu et~al.(2018)Yu, Guo, Yi, Chang, Potdar, Cheng, Tesauro, Wang, and
  Zhou}]{yu2018diverse}
Mo~Yu, Xiaoxiao Guo, Jinfeng Yi, Shiyu Chang, Saloni Potdar, Yu~Cheng, Gerald
  Tesauro, Haoyu Wang, and Bowen Zhou. 2018.
\newblock Diverse few-shot text classification with multiple metrics.
\newblock \emph{arXiv preprint arXiv:1805.07513}.

\end{thebibliography}
\bibliographystyle{acl_natbib}

\appendix
\section{Appendix}
\begin{figure}[t!]
    \centering
    \includegraphics[width=\linewidth]{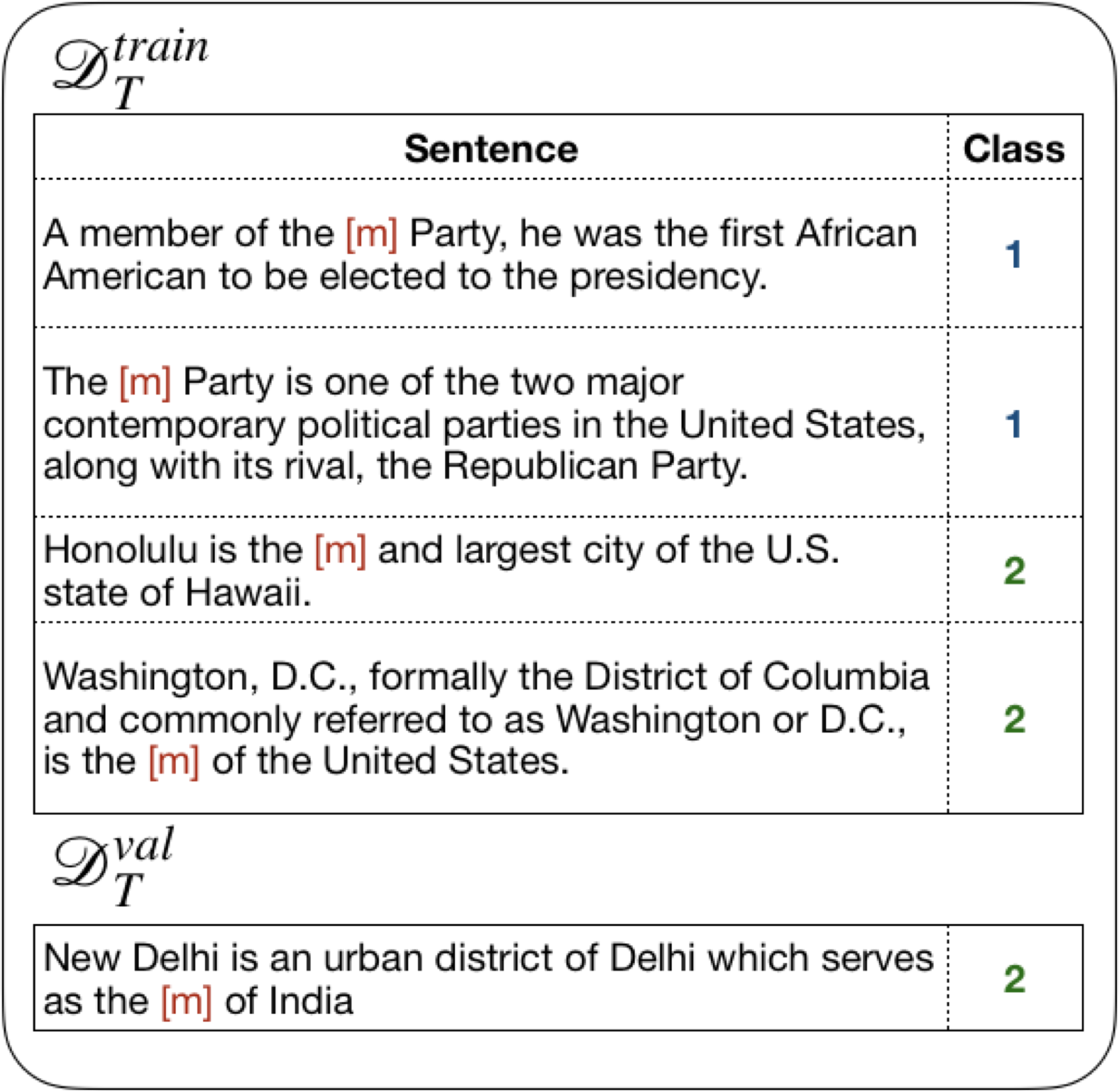}
    \caption{Example of a task in SMLMT.}
    \label{fig:smlmt}
\end{figure}
\begin{figure}[t!]
    \centering
     \includegraphics[width=\linewidth]{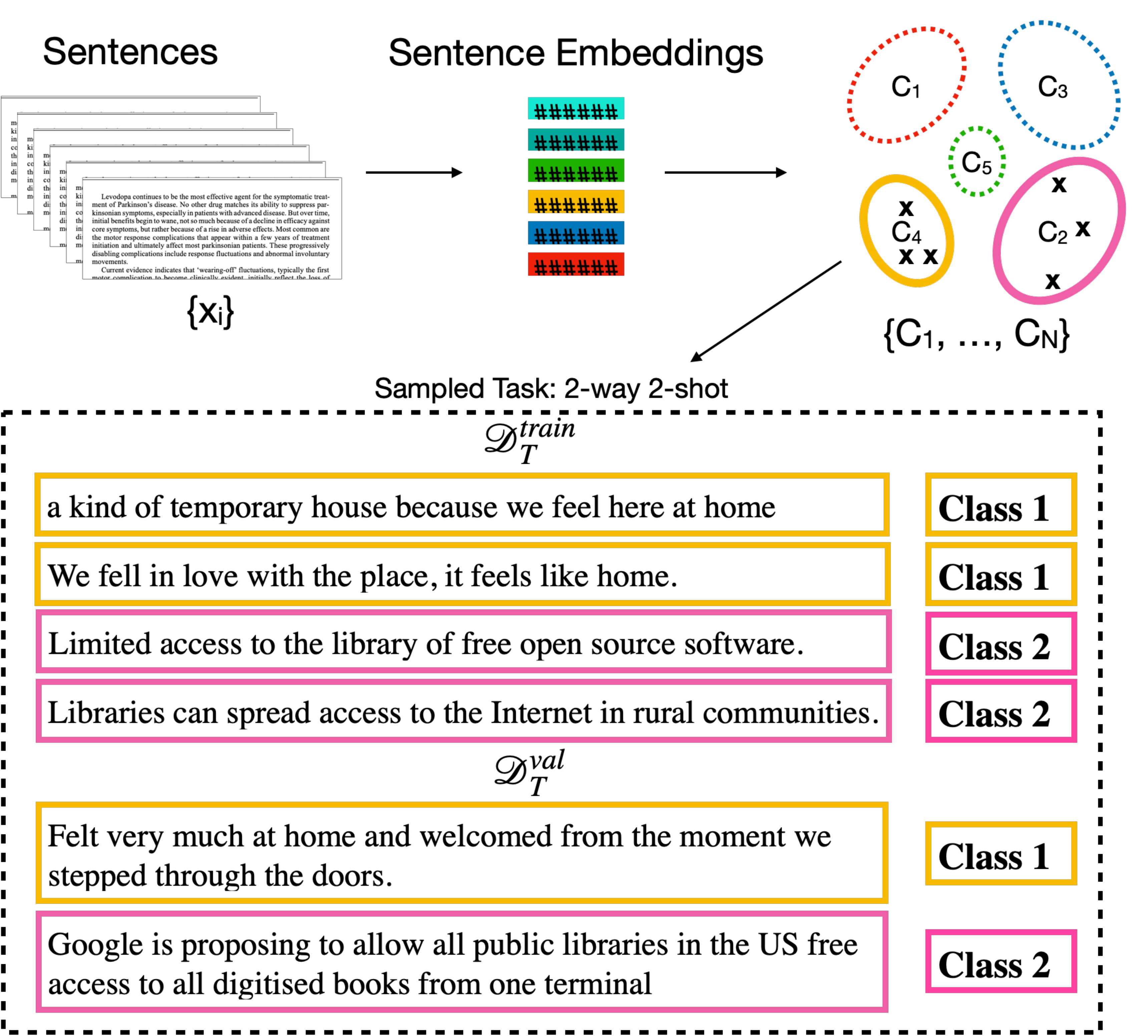}
    \caption{Illustration of SentCluster approach.}
    \label{fig:sentcluster}
\end{figure}

Examples of SMLMT and SentCluster can be seen in Figures~\ref{fig:smlmt} and \ref{fig:sentcluster}.

\begin{table}[tb!]
\centering \fontsize{8.0}{9.5}\selectfont \setlength{\tabcolsep}{0.5em}
\begin{tabular}{lcccc}
\Xhline{2\arrayrulewidth}\\[-1.3\medskipamount]
    \multirow{2}{*}{Model} & \multicolumn{2}{c}{1-shot} & \multicolumn{2}{c}{5-shot} \\
     & $N=5$ & $N=10$ & $N=5$ & $N=10$\\[3pt] \Xhline{2\arrayrulewidth}\\[-1.3\medskipamount]
    SentCluster & 
     34.2  & 21.2 & 52.3 & 36.3 \\
    Uniform &
     55.8  & 40.1 & 76.1 & 61.0 \\
    Frequency & 
     57.6  & 41.6 & \underline{78.1} & 61.5 \\
    Cluster & 
     \underline{60.4} & \underline{45.2}  & \underline{78.1} & \underline{63.9} \\
    Cluster-ccnet & 
     \textbf{60.1} & \textbf{46.0} & \textbf{81.2} & \textbf{67.6} \\ \midrule
\bottomrule
\end{tabular}
\caption{Results on Fewrel 2.0 validation.}
\label{tab:fewrel_val}
\end{table}

\subsection{Additional Experiment Results}
\begin{figure*}[t!]
    \centering
    \includegraphics[width=\linewidth]{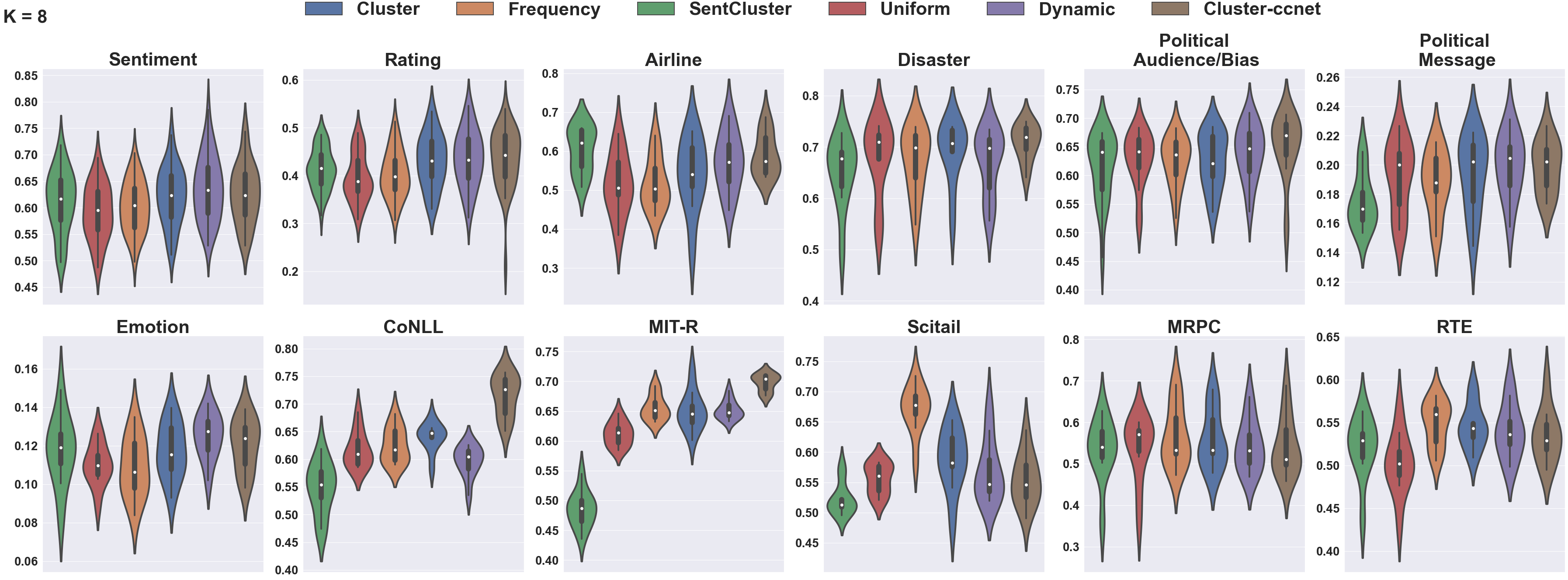}
    \includegraphics[width=\linewidth]{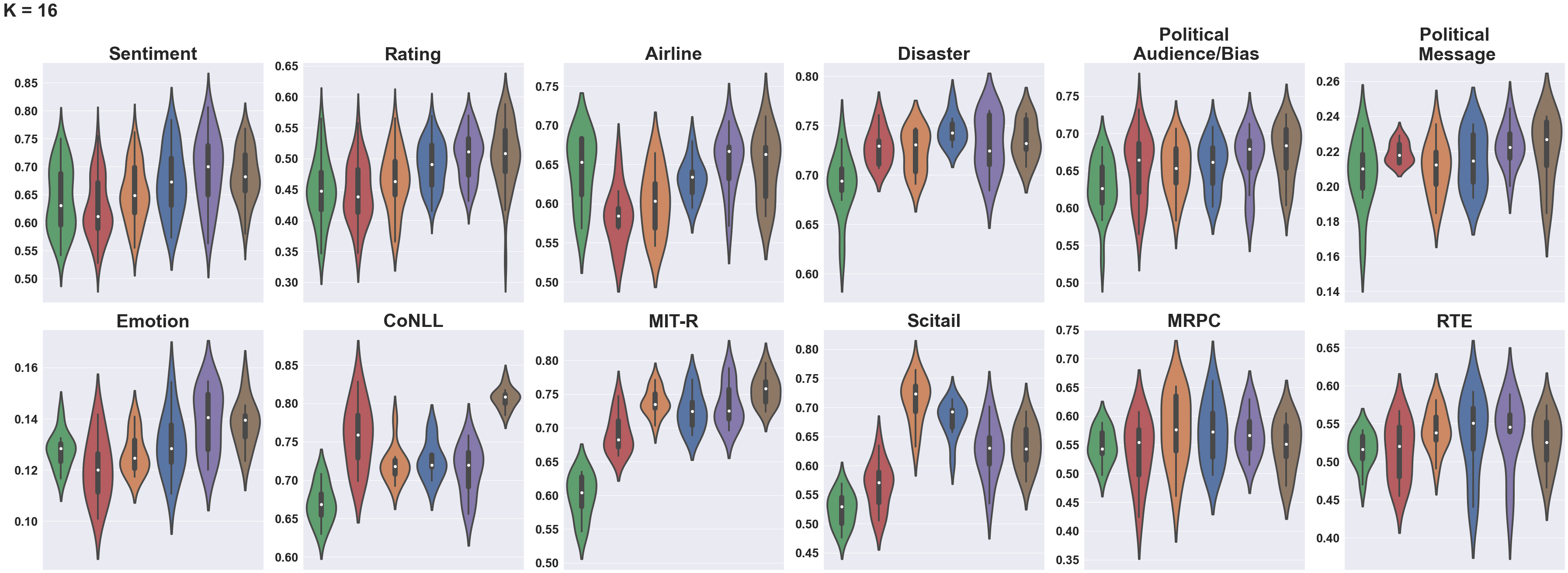}
    \includegraphics[width=\linewidth]{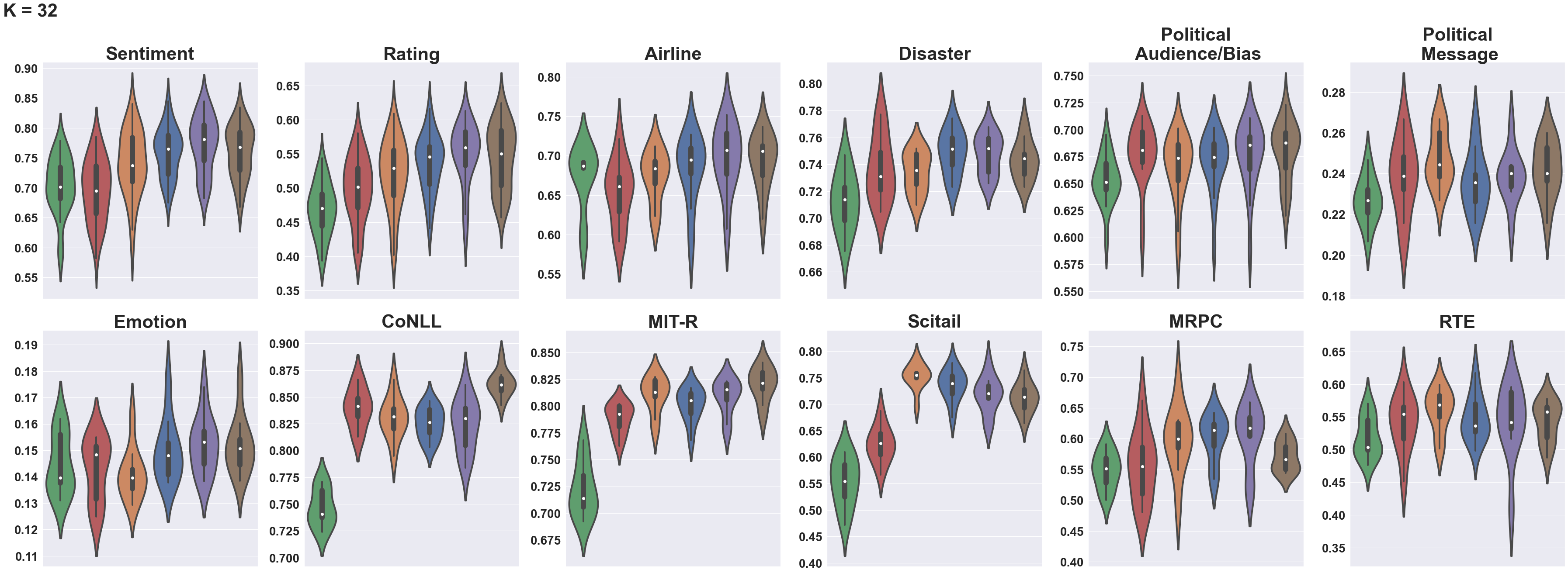}
    \caption{Results across all tasks. Sentiment and Rating are average of 4 domains used in \cite{bansal2020learning}.Each violin plot for a model shows the full distribution of accuracy across multiple runs (and domains).}
    \label{fig:violin}
\end{figure*}

Results on FewRel 2.0 validation set using the different task distributions is shown in Figure \ref{tab:fewrel_val}.
Full distribution of results in down-stream takss for the various self-supervised tasks can be seen in Fig.~\ref{fig:violin}

\subsection{Fine-tuning hyper-parameters} 
\begin{table}[t]
\centering
\resizebox{\linewidth}{!}{%
\begin{tabular}{cc}
\Xhline{2\arrayrulewidth}
Hyper-parameter & Value \\ \Xhline{2\arrayrulewidth}
Tasks per batch & 4 \\ 
$d$ & 256  \\ 
Attention dropout & 0.1 \\ 
Hidden Layer Dropout & 0.1\\ 
Outer Loop Learning Rate & 1e-05\\ 
Adaptation Steps ($G$) & 7\\ 
Meta-training Epochs & 1 \\
Lowercase text & False \\ 
Sequence Length & 128 \\ 
Learning-rate Warmup & 10\% of steps \\ \midrule
SentPair ratio $\alpha$ & $\frac{1}{16}$ \\
Number of Tasks & 4 Million \\
Support samples per task & 80\\ 
Query samples per task & 10\\ 
Number of classes for tasks & [2,3,4,5] \\ 
Number of Clusters $M$ & 500 \\
Number of Clusters in SentCluster & 200k \\
$\lambda_t$ in Dynamic & Anneal \\
$m$ interval in Dynamic & 5000 \\
\Xhline{2\arrayrulewidth}
\end{tabular}
}
\caption{Hyper-parameters. The parameters relating to the task distributions are in the bottom section of the table.}
\label{tab:hyper}
\end{table}
The meta-learning methods learn the learn rate for fine-tuning, thus we only tune the number of steps to run fine-tuning by using development data from 2 tasks (Scitail, Amazon Electronics), following \citet{bansal2020learning,bansal2020self}.
We found that running fine-tuning until the loss on the support set is small ($<=0.01$) is an alternative that also performs competitively and does not require tuning the number of steps.
The reported results followed the previous approach and the tuned number of steps of fine-tuning for $k=8,16,32$ respectively were: (1) Uniform: 100,75,100 (2) Frequency: 25,150,75 (3) Cluster: 75,50,75 (4) Cluster-ccnet: 150,200,75 (5) SentCluster: 100,250,25 (6) Dynamic: 10, 100, 200. On FewRel we found 20 steps of updates on the support set to perform well on the validation data for all settings.

\subsection{Other Implementation Details} 
Since the Fewrel tasks consist of entity pair in the sentence it is important to mark these entities which define the relation to be classified. We used unused tokens in the BERT vocabulary to mark the positions of the entity mentions. Note the in the unsupervised models these unsused tokens get a zero-embedding and are only fine-tuned from the 1-shot or 5-shot support sets.

Hyper-parameters for meta-training are listed in Table \ref{tab:hyper}. Dataset statistics for downstream classification tasks can be found in \citet{bansal2020learning} and few-shot splits can be downloaded from \url{https://github.com/iesl/leopard}.

Training Hardware:
The models were trained on 32 V100 GPU. Training takes about 42 hours.





\end{document}